\documentclass[format=sigconf]{acmart}

\usepackage[utf8]{inputenc}
\usepackage{eqnarray}
\usepackage{algorithm}
\usepackage{enumitem}
\usepackage[noend]{algpseudocode}
\usepackage{amsmath}
\usepackage{xspace}
\usepackage{mathtools}
\usepackage{appendix}
\usepackage{microtype}

\newcommand{\mome}{\textsc{mome}\xspace}
\newcommand{\pgx}{\textsc{\mome -pgx}\xspace}

\newcommand{\pga}{\textsc{pga-me}\xspace}
\newcommand{\spea}{\textsc{spea2}\xspace}
\newcommand{\nsga}{\textsc{nsga-ii}\xspace}

\newcommand{\mopga}{\textsc{mo-pga}\xspace}
\newcommand{\mopgaenergy}{\textsc{mo-pga (only energy)}\xspace}
\newcommand{\mopgaforward}{\textsc{mo-pga (only forward)}\xspace}

\newcommand{\momecrowding}{\textsc{mome-crowding}\xspace}

\newcommand{\mapelites}{\textsc{map-elites}\xspace}
\newcommand{\tdthree}{\textsc{td3}\xspace}

\newcommand{\cmame}{\textsc{cma-me}\xspace}
\newcommand{\mcx}{\textsc{mcx}\xspace}
\newcommand{\dqd}{\textsc{dqd}\xspace}
\newcommand{\memap}{\textsc{me-}\mapelites}

\newcommand{\moqd}{\textsc{moqd}\xspace}

\newcommand{\qd}{\textsc{qd}\xspace}
\newcommand{\mo}{\textsc{mo}\xspace}
\newcommand{\mdp}{\textsc{MDP}\xspace}
\newcommand{\mdps}{\textsc{MDPs}\xspace}

\newcommand{\ga}{\textsc{ga}\xspace}
\newcommand{\pg}{\textsc{pg}\xspace}

\newcommand{\replications}{\textsc{15}\xspace}

\newcommand{\initialisation}{// \texttt{Initialisation}\xspace}
\newcommand{\mainloop}{// \texttt{Main loop}\xspace}
\newcommand{\sample}{// \texttt{Sample solutions}\xspace}
\newcommand{\generate}{// \texttt{Generate offspring}\xspace}
\newcommand{\offspringeval}{// \texttt{Evaluate offspring}\xspace}

\newcommand{\archiveadd}{// \texttt{Add to archive}\xspace}
\newcommand{\updateiter}{// \texttt{Update iterations}\xspace}

\newcommand{\qdscore}{\textsc{qd-score}\xspace}
\newcommand{\moqdscore}{\textsc{moqd-score}\xspace}
\newcommand{\maxsumscores}{\textsc{maximum sum of scores}\xspace}
\newcommand{\globalhypscore}{\textsc{global-hypervolume}\xspace}
\newcommand{\coverage}{\textsc{coverage}\xspace}

\newcommand{\moqdpvalue}{\textsc{$2.2\times10^{-2}$}\xspace}
\newcommand{\globalhyppvalue}{\textsc{$4.3\times10^{-3}$}\xspace}
\newcommand{\maxscorespvalue}{\textsc{$4.8\times10^{-4}$}\xspace}

\newcommand{\psumscoresant}{\textsc{$5\times10^{-4}$}\xspace}

\newcommand{\pscoreablations}{\textsc{$1\times10^{-2}$}\xspace}

\newcommand{\code}[0]{\url{https://github.com/adaptive-intelligent-robotics/MOME_PGX}}

\begin{document}

\acmConference[GECCO '23]{Genetic and Evolutionary Computation Conference}{July 15--19, 2023}{Lisbon, Portugal}
\acmDOI{10.1145/3583131.3590470}
\acmISBN{979-8-4007-0119-1/23/07}
\acmYear{2023} 
\copyrightyear{2023}

\title{Improving the Data Efficiency of Multi-Objective Quality-Diversity through Gradient Assistance and Crowding Exploration}

\renewcommand{\shorttitle}{Improving the Data Efficiency of Multi-Objective Quality-Diversity}

\author{Hannah Janmohamed}
\affiliation{%
  \institution{Imperial College London, InstaDeep}
  \city{London}
  \country{United Kingdom}}
\email{hnj21@imperial.ac.uk}

\author{Thomas Pierrot}
\affiliation{%
 \institution{InstaDeep}
  \city{Boston}
  \country{USA}}
\email{t.pierrot@instadeep.com}

\author{Antoine Cully}
\affiliation{%
  \institution{Imperial College London}
  \city{London}
  \country{United Kingdom}}
\email{a.cully@imperial.ac.uk}

\renewcommand{\shortauthors}{Janmohamed, et al.}

\begin{abstract}

Quality-Diversity (\qd) algorithms have recently gained traction as optimisation methods due to their effectiveness at escaping local optima and capability of generating wide-ranging and high-performing solutions.
Recently, Multi-Objective MAP-Elites (\mome) extended the \qd paradigm to the multi-objective setting by maintaining a Pareto front in each cell of a \mapelites grid. 
\mome achieved a global performance that competed with \nsga and \spea, two well-established multi-objective evolutionary algorithms, while also acquiring a diverse repertoire of solutions. 
However, \mome is limited by non-directed genetic search mechanisms which struggle in high-dimensional search spaces. 
In this work, we present Multi-Objective MAP-Elites with Policy-Gradient Assistance and Crowding-based Exploration (\pgx): a new \qd algorithm that extends \mome to improve its data efficiency and performance. 
\pgx uses gradient-based optimisation to efficiently drive solutions towards higher performance. 
It also introduces crowding-based mechanisms to create an improved exploration strategy and to encourage greater uniformity across Pareto fronts.
We evaluate \pgx in four simulated robot locomotion tasks and demonstrate that it converges faster and to a higher performance than all other baselines.
We show that \pgx is between 4.3 and 42 times more data-efficient than \mome and doubles the performance of \mome, \nsga and \spea in challenging environments.

\end{abstract}



\maketitle

\begin{figure}[ht!]
  \centering
  \includegraphics[width=0.42\textwidth]{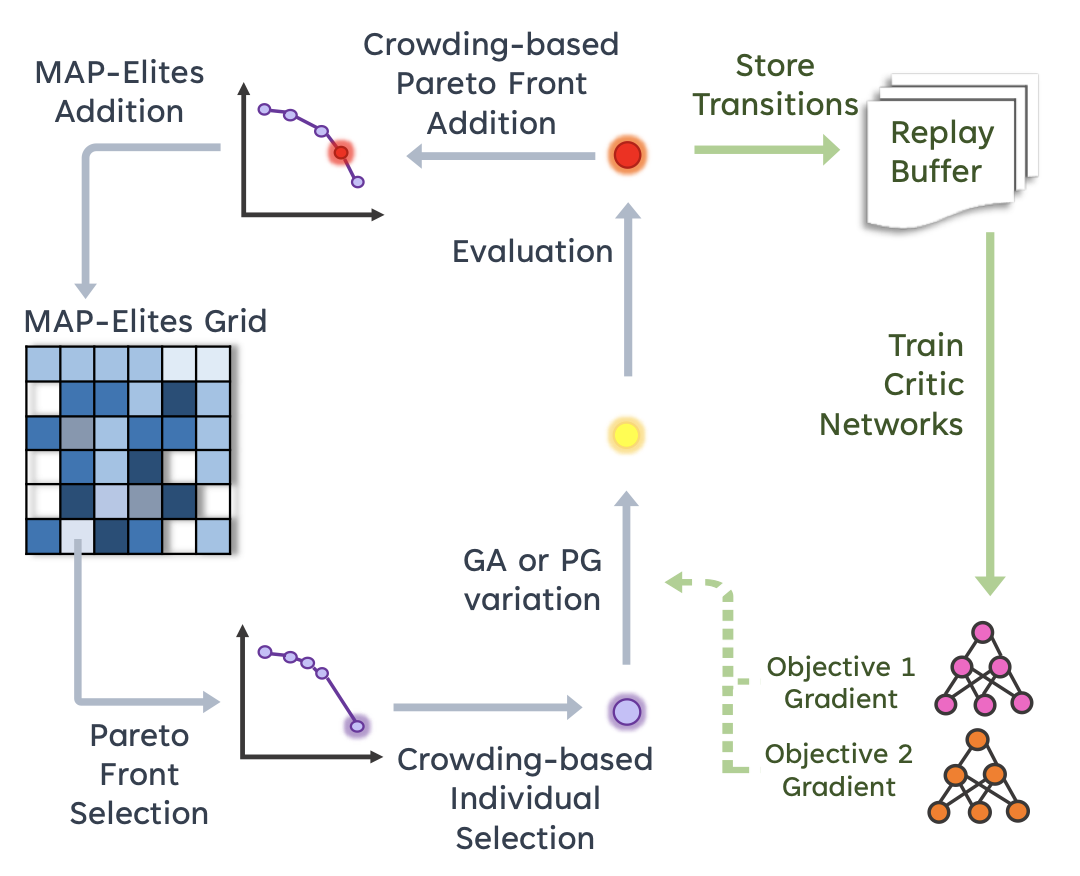}
  \caption{\pgx algorithm. At each iteration, solutions are sampled, mutated and added back to the archive based on their Pareto-dominance and crowding distance. Separate to this main \qd loop, a critic network per objective function is trained periodically, using transitions from a buffer.}
  \label{fig:teaser}
\end{figure}

\vspace{-10pt}
\section{Introduction}

Quality-Diversity (\qd) algorithms form a new optimisation paradigm designed to acquire a set of high-performing and diverse solutions, rather than the single optimum obtained by many traditional optimisation methods. 
This approach has proven useful in many domains where having more than one solution is a desirable outcome. 
For example, in robotics, learning a repertoire of controllers allows the robot to adapt in the event of damage \cite{nature, hbr}. 
Similarly, \qd algorithms have also been used to generate portfolios of architectural designs \cite{mcx}, create video game content \cite{videogames} and manage scheduling tasks \cite{scheduling}. 
Even in applications where only a single solution is sought, \qd has been observed to be a highly effective optimisation approach \cite{qdsteppingstones}.
Specifically, maintaining a divergent population of solutions creates an implicit exploration strategy, helping the optimisation process to avoid getting stuck in local minima and resulting in a higher global performance \cite{noveltysearch, qdsteppingstones}.

As \qd optimisation becomes increasingly capable of tackling complex problems, attention turns to how we can leverage them in the multi-objective case. 
Indeed, many real-life applications don't involve just a single objective, but rather a multitude of sophisticated and potentially contrasting ones. 
Canonical examples, such as wind turbine control \cite{hayes2022practical} or computer hardware design \cite{spea}, often seek the maximisation of some performance objective whilst minimising a cost function and obeying other safety constraints.
But the scope and range of multi-objective problems is vast, including domains that range from drug design \cite{mo_molecular} to satellite communication \cite{mo_satellites}.

The recently introduced Multi-Objective MAP-Elites (\mome) \cite{mome} unifies methods from  Multi-Objective (\mo) optimisation with the \qd framework. 
Traditional \mo algorithms don't aim to find a single optimal solution, but rather a set of solutions, each of which represents a different optimal trade-off of the given objectives. 
This set, referred to as the Pareto front, provides the end user with a variety of solutions to pick from. 
To incorporate this goal into \qd, \mome aims to find a large collection of Pareto fronts, each of which occupies a different behavioural niche. 
While \mome shows promising results, it relies on non-directed genetic mutation operators, making it limited to simple tasks with low-dimensional search spaces.

In this paper, we present Multi-Objective MAP-Elites with Policy-Gradient Assistance and Crowding-based Exploration (\pgx): a data-efficient and high-performing Multi-objective Quality-Diversity (\moqd) algorithm for tasks that can be formulated as Markov Decision Processes (\mdp).
The algorithm is visualised in Figure \ref{fig:teaser} and builds on top of \mome: Pareto fronts of solutions are stored in cells of a \mapelites grid and, at each iteration, solutions are selected, mutated and potentially re-added to the grid based on their fitness. 
However, our proposed method makes two key contributions to improve the performance and data efficiency of \mome. 
Firstly, \pgx uses gradient-based mutations to find high performing solutions for each objective.
Secondly, it uses crowding-based criteria to alter the selection and addition mechanisms of \mome in order to increase exploration in sparse regions of the objective space and encourage a uniform distribution of points on the Pareto fronts.

We evaluate \pgx on four simulated robotic control tasks using large neural-networks and compare its performance to a variety of standard \qd and \mo baselines. 
We find that \pgx outperforms all standard baselines across all \moqd metrics in each of the tasks and doubles the performance of \mome, \nsga and \spea in challenging tasks.
Furthermore, \pgx is shown to be between 4.3 and 42 times more data-efficient than \mome.
Finally, we provide an ablation study to demonstrate that encompassing gradients and using crowding-mechanisms are both essential for improving the performance of \mome. 
Our code implementation was containerised and implemented using the QDax framework \cite{qdax} for all methods and baselines, and is available at: \code.

\section{Background}


\subsection{Quality-Diversity Optimisation}

Traditional optimisation problems aim to find the single solution $x$ from a search space $\mathcal{X}$ which maximises an objective function $f : \mathcal{X} \to \mathbb{R} $. 
By contrast, Quality-Diversity (\qd) algorithms extend this objective by additionally encompassing a descriptor function $c: \mathcal{X} \to \mathbb{R}^d$ which maps solutions from the search space to $d$-dimensional descriptor vectors. 
Rather than seek a single solution $x$, \qd algorithms aim to find a population of solutions $x_i \in X$ which both maximise the objective $f$ and have diverse descriptor vectors.

\mapelites algorithms \cite{mapelites, mome, cvt} are a simple, yet powerful family of \qd algorithms which achieve the \qd goal by discretising the set of possible behaviour descriptors $S = c(\mathcal{X})$, referred to as the descriptor space, into a grid structure with $k$ cells $S_i$ which is used to store solutions. 
Specifically, the solution $x_i$ is stored in the cell $S_i$ that corresponds to its descriptor $c(x_i)$.
In each iteration of \mapelites, a batch of solutions is selected from the grid and mutated to form new solutions. 
Then, the new solutions are evaluated and potentially added back to the grid structure. 
Specifically, if the cell corresponding to the new solution's descriptor is empty, it is added.
Otherwise, the new solution only replaces the existing one if it is higher-performing.
This loop is repeated for a set budget and, as the algorithm progresses, the grid structure becomes incrementally populated with more diverse and higher-quality solutions. 

 
\subsection{Multi-objective Optimisation}

A multi-objective optimisation problem considers the simultaneous maximisation of many objectives, $f_1(x), f_2(x), ... f_m(x)$, rather than a single one. 
Usually, these objectives come as a trade-off so there is not a single optimal solution, but rather a set of solutions that achieve different compromises across these objectives.
Accordingly, the notion of Pareto-dominance is often used to induce a preference over solutions. 
Given two solutions $x_1$ and $x_2$ we say that $x_1$ dominates $x_2$ \cite{spea}, or $x_1 \succ x_2$, if:

\vspace{-10pt}
\begin{multline}
    \forall i \in \{1, 2, ..., m\}: f_i(x_1) \geq f_i(x_2)\\
    \wedge 
    \\
   \exists j \in \{1, 2, ..., m\}: f_j(x_1) > f_j(x_2)\,.
\end{multline}

That is, one solution dominates another if it scores at least as high as the other across all objectives and strictly higher in at least one.
Given a set of candidate solutions $\mathcal{S}$, the Pareto front is the set of solutions $\mathcal{P}(\mathcal{S})\in\mathcal{S}$ that are not dominated by any other solutions in $\mathcal{S}$.
In other words, the Pareto front represents all of the best possible trade-offs of the objective functions.
The goal of \mo optimisation is to approximate the Pareto front over the entire search space of solutions, $\mathcal{P}(\mathcal{X})$, known as the optimal Pareto front.

\begin{figure}[ht]
  \centering
  \includegraphics[width=0.55\linewidth]{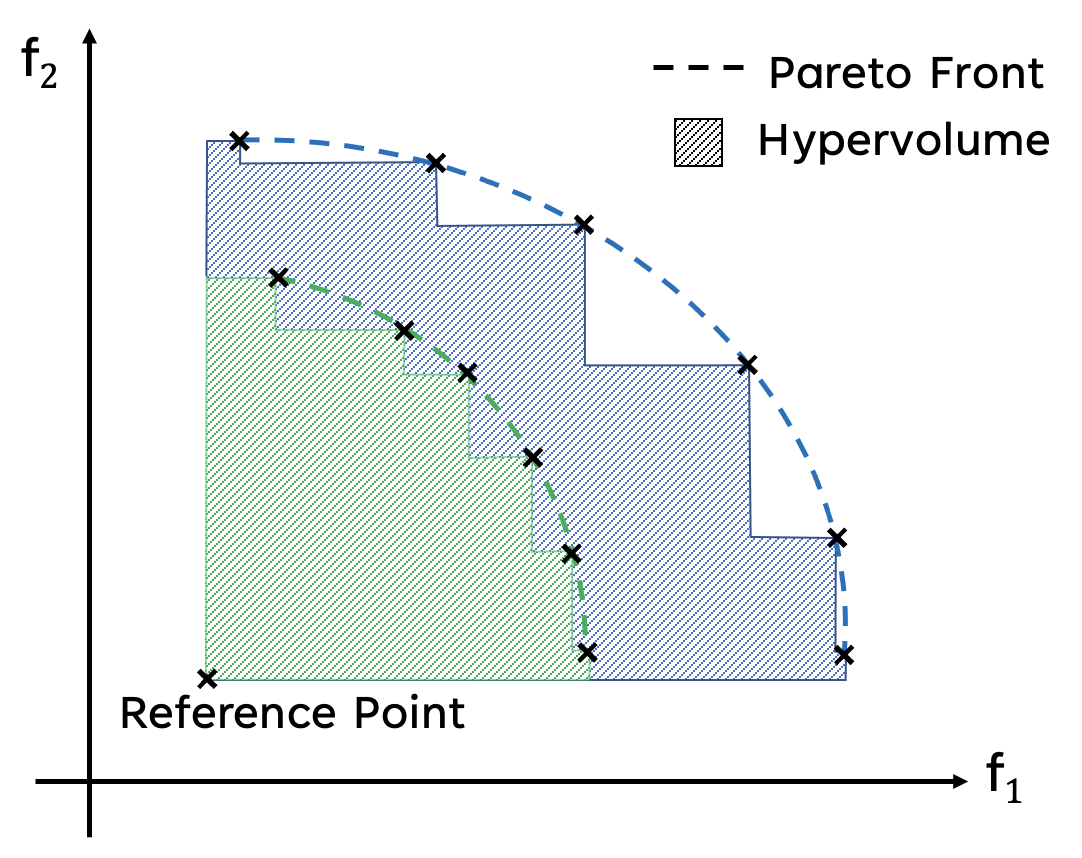}
  \caption{Two Pareto fronts and their hypervolumes with respect to a given reference point for a bi-objective problem. All of the points on the blue (outer) Pareto front strictly dominate the points on the green (inner) Pareto front and, accordingly, the blue Pareto front has a larger hypervolume.}
  \label{fig:pf}
\end{figure}

The performance of \mo algorithms is commonly quantified via the hypervolume metric \cite{moea_survey, usinghypervolumes} which reflects the coverage of the objective space achieved by solutions on the Pareto front. 
As visualised in Figure \ref{fig:pf}, for two objectives, this corresponds to the union of rectangles covered by solutions on the Pareto front \cite{zitzler1998multiobjective}. More generally, the hypervolume $\Xi$ of a Pareto front $\mathcal{P}$ is defined as:
\begin{equation}
    \Xi(\mathcal{P}) = \lambda ({x \in \mathcal{X} \,|\, \exists \, s \in \mathcal{P}, s \succ x \succ r})\,,
   \label{eqn:hypervolume}
\end{equation}

where $\lambda$ denotes the Lebesgue measure and $r$ is a pre-determined and fixed reference point \cite{usinghypervolumes, mome}. 
Limitations to this metric include its reliance on defining a suitable reference point $r$ which can impact on its magnitude. 
However, it is regarded as a reliable \mo performance metric due to its monotonic and scale-invariant properties \cite{usinghypervolumes}.





\section{Related Works}

\subsection{Multi-Objective Quality-Diversity}

Multi-Objective Quality-Diversity (\moqd) combines the goals of \qd and \mo optimisation, with the aim of returning the Pareto front with maximum hypervolume in each cell of the descriptor space.
To the best of our knowledge, Multi-Objective MAP-Elites (\mome) \cite{mome} is the only proposed method that has been specifically designed to tackle the \moqd problem. 
Its algorithmic approach is a natural extension to \mapelites where, in each iteration of the algorithm, solutions are randomly selected from the population, mutated and potentially re-added to the population based on their fitness. 
However, unlike \mapelites which stores at most one solution in each cell, \mome stores a Pareto front in each descriptor cell.
If a new solution is on the Pareto front of the cell that corresponds to its descriptor, it is added to the grid. 
Otherwise, if the new solution dominates an existing solution on this front, it replaces it. 
Since the new solution could dominate several solutions in the cell's existing Pareto front, it is possible for it to replace more than one solution.
\mome was shown to achieve a competitive \mo performance compared to popular \mo baselines, while also acquiring a diverse repertoire of behaviours.
 
Another method of note is Multi-Criteria Exploration (\mcx) \cite{mcx} which also follows a usual \mapelites loop but uses a tournament ranking strategy to assess a solution's performance across multiple objectives.
Specifically, \mcx uses a "T-DominO score" which reflects how balanced solutions are across all objectives. 
If new a solution achieves a higher T-DominO score than the existing solution in the cell, it replaces it.
While this approach is desirable in some contexts, it makes the assumption a priori that all objectives are of equal preference. 
Furthermore, \mcx does not aim to obtain a Pareto front of solutions and so, while we note its relevance, we consider it to be closer related to traditional mono-objective \qd methods.

\subsection{Data-efficient Quality-Diversity}
Recent works have proposed modifications to \qd algorithms in order to improve their computational cost. 
For example, some methods use surrogate models to estimate the performance of solutions without directly evaluating them \cite{sail, mbqd, daqd, rfqd}. 
As such, solutions that are predicted to be low-performing can be filtered out and unnecessary computation can be avoided. 
While this can reduce the number of required evaluations by more than an order of magnitude \cite{daqd}, it relies on having a accurate surrogate model which may be difficult to obtain.

Other works focus on creating a better exploration strategy. 
For example, Covariance Matrix Adaptation MAP-Elites (\cmame) \cite{cma} uses several emitter distributions to guide the exploration process toward the direction of, for example, highest diversity or performance. 
Or, more recently, Multi-Emitter MAP-Elites (\memap) includes a heterogeneous set of emitters and then uses a bandit algorithm to determine which combination of them to use to maximise the proportion of solutions that are added to the archive.


Some \qd methods only consider the subset of problems that can be formalised as \mdps.
In these tasks, solutions $x_i\in\mathcal{X}$ actually encode parameters for policies $\pi_{x_i}$, for an agent that interacts in a sequential environment at discrete time-steps $t$. 
At each time step, the agent observes its current state $s_t$ and chooses an action $a_t$ which leads to a subsequent state $s_{t+1}$ and reward $r_t$. The policy of the agent determines which action is taken in each time-step, $\mathbb{P}(a_t|s_t)$, and the aim of the task is to find the policy that maximises the expected return of the agent over $T$ time-steps:

\vspace{-5pt}
\begin{equation}
    J(x) = \sum_{j=0}^T \gamma^j r(s_j, a_j)\,.
\end{equation}

 Here, $\gamma$ is known as a discount factor: a weighting coefficient that expresses the preferences between short-term and future rewards.


Policy Gradient Assisted MAP-Elites (\pga) takes tasks from this \mdp setting and employs techniques from the \tdthree algorithm \cite{td3} in order to improve the optimisation process.
Specifically, as solutions are evaluated in the main \qd loop, transition tuples of $(s_t, a_t, r_t, s_{t+1})$ are collected and stored in a replay buffer $\mathcal{B}$. 
These samples are used to train a critic network $Q_\theta$ that approximates the action-value function,

\vspace{-5pt}
\begin{equation}
    Q(s_t, a_t) = \mathbb{E} \bigg[\sum_{j=t}^T \gamma^j r(s_j, a_j)\bigg]\,,
\end{equation}  

which captures the value of taking action $a_t$ from the state $s_t$. These critic estimates can consequently be used to form a policy gradient estimate:

\vspace{-5pt}
\begin{equation}
    \nabla _{x_i} J(x_i) = \mathbb{E}\big[\,\nabla _{x_i}\,\pi _{x_i}(s)\,\nabla _a Q_\theta (s, a)\,|\,_{a=\pi _{x_i}(s)} \big]\,.
\end{equation}  

Intuitively, the policy gradient provides the direction in which a solution $x_i$ should be updated in order to maximise the reward of the agent. 
In each iteration of \pga some solutions are mutated via gradient-based updates and, to maintain exploration, others are mutated via genetic mutations. This accelerates the optimisation process since gradient-information quickly highlights promising regions of the solution space, while genetic search enforces thorough exploration in these regions \cite{pgaempirical}. 
 
More recently, Pierrot and Macé et al. \cite{qdpg} built upon \pga by using gradients to improve both the fitness and the diversity of solutions. Specifically, rather than rely on genetic variations for exploration, they introduced a Diversity Policy Gradient operator to improve exploration of states at each time step.
Alternatively, Differentiable Quality-Diversity (\dqd) \cite{dqd} is a gradient-based approach that does not rely on an \mdp formulation but assumes differentiable objective and descriptor functions, and incorporates their gradients into a variation operator which greatly improves search efficiency. 
Subsequently, Tjanaka et al. adapted this algorithm for \mdp tasks where the environment is not differentiable by instead using evolutionary strategies and techniques from \tdthree to form gradient approximations \cite{dqdformdps}.

\subsection{Multi-objective Evolutionary Algorithms}

Evolutionary algorithms are often applied to \mo problems since their population-based approach lends itself naturally to finding Pareto fronts. 
In general, they work by initialising a population of solutions and then employing carefully designed selection, mutation and replacement operators in order to encourage exploration in promising regions of the solution space and hence drive performance of solutions toward the optimal Pareto front.

\nsga \cite{nsga2} is one such method that is driven by biased selection mechanisms. At each iteration, the population is first organised into sets of Pareto fronts: the first front is the Pareto front over all solutions; the second front is the Pareto front over all solutions except those that belong to the first front, and so on. Then solutions in each front are assigned a crowding distance score which reflects the density of the region of objective space that they occupy. Finally, an ordering over all solutions is calculated based on the previous two steps: solutions belonging to higher fronts and lying in less dense areas of the objective space are preferred. 
This ordering is used to bias the selection probability of solutions.
As a result, \nsga has been demonstrated to achieve Pareto fronts that are both high performing and well spread.  

\spea \cite{spea2} is an alternative, long-established \mo algorithm that has a similar approach. Specifically, each individual is assigned a strength score that reflects the number of solutions that it dominates. 
Then, each solution is assigned a raw strength equal to the total strength scores of the solutions that dominate it. A lower raw strength is desirable as it means that it is dominated by fewer solutions. The selection probability of a solution is proportional to its raw strength and the corresponding density of the objective space that it lies in. 
Similar to \nsga, this encourages more exploration around individuals that are high-performing and that lie in less populated regions of the objective space.

A multitude of other \mo evolutionary algorithms exist and we refer the reader to a comprehensive survey \cite{moea_survey} for a summary of other approaches. However, to the best of our knowledge, \mome is the only \mo method that actively maintains diversity over the descriptor space.

\section{MOME-PGX}

In this section, we present \pgx, a new \moqd algorithm. 
Our proposed method uses policy gradient (\pg) mutation operators in order to efficiently drive solutions towards high performance in each objective, while maintaining a divergent search through traditional genetic variation operators. 
\pgx also improves on the uniform selection and replacement mechanisms from \mome by instead encouraging exploration around solutions that lie in less crowded regions of the objective space.
Details for each of these contributions are found in the next sections and the pseudo-code for the \pgx can be found in Algorithm \ref{alg:pseudo}.

\begin{algorithm}
\caption{\pgx pseudo-code}\label{alg:pseudo}
\begin{algorithmic}

\State{\textbf{Input:}}
\begin{itemize}
    \item A \mapelites archive $\mathcal{A}$, with $k$ cells, and an empty Pareto front of size $P$ in each cell
    \item the batch size $B$ and total number of iterations $N$
    \item $m$ objective functions $f_1, ..., f_m$ and a descriptor function $c$
    \item an empty replay buffer $\mathcal{B}$
    \item pairs of actor, critic and target networks for $f_1, ..., f_m$
\end{itemize}

\\
\State{\initialisation}
\State{Generate initial candidate solutions $x_k$}
\State{Find descriptors $c(x_k)$ and fitnesses $f_1(x_k), ..., f_m(x_k)$ of $x_k$}
\State{Add initial solutions to $\mathcal{A}$ and update Pareto fronts}
\State{Add transitions collected by $x_k$ to buffer $\mathcal{B}$}
\State{Train actor and critic networks on samples from $\mathcal{B}$}
\\

\State{\mainloop}
\For{$\text{iter = 1} \to N $}
    
    \State{\sample}
    \State{Sample $B$ non-empty cells from $\mathcal{A}$ with uniform probability}
    \State{Calculate crowding distances in each sampled cell}
    \State{In each sampled cell's Pareto front, sample a solution $x$ with probability proportional to the crowding distances of solutions}

    \\
    \State{\generate}
    \State{Apply \ga variation to $B/2$ sampled solutions} 
    \State{Apply \pg variation to $(B/2)\div m$ sampled solutions for $m$ objective functions} 

    \\
    \State{\offspringeval}
    \State{Find descriptors and fitness values for offspring}
    \State{Store transitions from offspring in buffer $\mathcal{B}$}
    \State{Train \pg networks}
    
    \\
    \State{\archiveadd}
    \State{Find cell corresponding to descriptor of each offspring}
    \State{Add offspring to cells and recompute Pareto fronts}
    \If{Pareto front length $>$ P}
        \State{Calculate crowding distances of solutions in cell}
        \State{Remove solutions with minimum crowding distance}
    \EndIf
    
    \\
    \State{\updateiter}
    \State{iter $\gets$ iter $+1$}
    \\

\EndFor
\Return{ $\mathcal{A}$}
\end{algorithmic}
\end{algorithm}

\subsection{Gradient-Assisted Mutations}


In \pgx, we extend the benefits of gradient-based optimisation that have been observed in \pga to the multi-objective case by maintaining a critic network per objective function. 
In each iteration, half of the solutions are mutated via a Genetic Algorithm (\ga) variation operator and the remaining solutions are divided into equal batch-sizes to be mutated with policy-gradients per objective function. 
So, for example, given a total batch-size of 256 and a bi-objective task, 128 of the solutions are mutated via the \ga variation operator, 64 are mutated according to the gradient for the first objective and the remaining 64 solutions are mutated according to the gradient of the second objective. \pgx also employs the same network training procedures as the \tdthree and \pga algorithms, such as using an actor network $\pi_\phi$ to calculate a target value for critic training and introducing pairs of critic networks $Q_{\theta_1}, Q_{\theta_2}$ to minimise bootstrapping errors. 
We refer the interested reader to the original \pga \cite{pgamap} and \tdthree \cite{td3} papers for further implementation details.

\subsection{Crowding-based selection and replacement}

In \pgx, we also introduce crowding-based selection and replacement mechanisms in order to induce a more targeted exploration strategy and improve the quality of the Pareto fronts obtained in each cell of the \mapelites grid.

To select offspring solutions at each iteration, \mome first uniformly samples a cell from the \mapelites grid and then uniformly samples a solution from the corresponding Pareto front.
To illustrate the limitations of this, we first note that since the objective functions are continuous, it is possible to fill gaps in a Pareto front with solutions that represent marginally different trade-offs, making the existing Pareto front more dense rather than actively extending its coverage.
Therefore, if solutions are not well spread on the Pareto front, uniform selection mechanisms can create an unintentional cycle: exploration is more likely to take place in more crowded areas of the Pareto front since there are more solutions there, which in turn may make that region even more dense, and so on. 
As this cycle continues, solutions that lie in sparser regions of the objective space will become increasingly less likely to be picked, and exploration in these regions will be neglected.

\begin{figure}[h]
  \centering
  \includegraphics[width=0.8\linewidth]{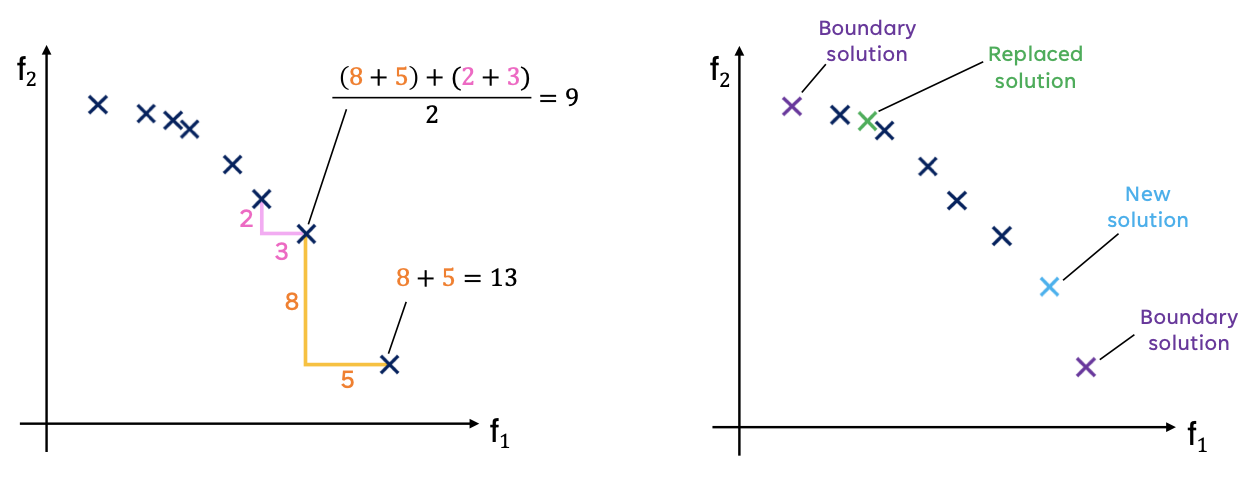}
  \caption{Left: crowding distances calculations for creating selection probabilities. Right: illustration of \pgx replacement mechanism. When the Pareto Front is at maximum capacity, the new solution replaces the solution which has the smallest crowding distance.}
  \label{fig:crowding}
\end{figure}

To address this, \pgx uses a biased selection mechanism based on a solution's crowding distance, similar to \nsga \cite{nsga2}.
As illustrated in Figure \ref{fig:crowding}, the crowding distance is calculated as the average Manhattan distance between solutions and their neighbours in the objective space \cite{nsga2}. 
For most solutions, we take the average distance to their two neighbouring solutions. However, for solutions at the boundaries of the Pareto front, we use the single nearest-neighbour.
When sampling solutions in each iteration of \pgx: first, a cell is chosen with uniform probability and then a solution from the corresponding Pareto front is selected with probability proportional to its crowding distance. By consequence, solutions in sparser regions of the front are more likely to be selected.

Another point of difference between \mome and \pgx is their Pareto front addition mechanisms. 
In both algorithms, the Pareto front in each cell is chosen to be a fixed maximum size (e.g., 50), which allows the algorithm to be parallelised and enables us to exploit the recent advances in massive hardware acceleration \cite{qdax}. 
However, in \mome, if a new solution is added to a cell that is at full capacity according to this maximum, it randomly replaces an existing solution. 
This approach pays no attention to the value of different solutions in terms of the trade-off of objectives they provide, and may result in loss of objective space coverage if boundary solutions are replaced. 
In particular, we note that two solutions that lie close together in the objective space provide the end user with a more similar trade-off of objectives than two solutions that lie farther apart. 
Therefore, we should prefer to retain solutions that are dissimilar in the objective space, to provide the end-user with the greatest variety and smoothness of contiguous trade-offs. 

\pgx addresses this limitation by retaining solutions according to their crowding distances. 
As illustrated in Figure \ref{fig:crowding}, if the Pareto front exceeds the maximum length, the solutions with the maximum crowding distances are retained.
In order to prevent erosion of coverage of the front, boundary points are always assigned infinite crowding distances, as in \spea \cite{spea2}.

\section{Experiments}

\subsection{Evaluation Tasks}
\begin{figure}[h]
  \centering
  \includegraphics[width=0.8\linewidth]{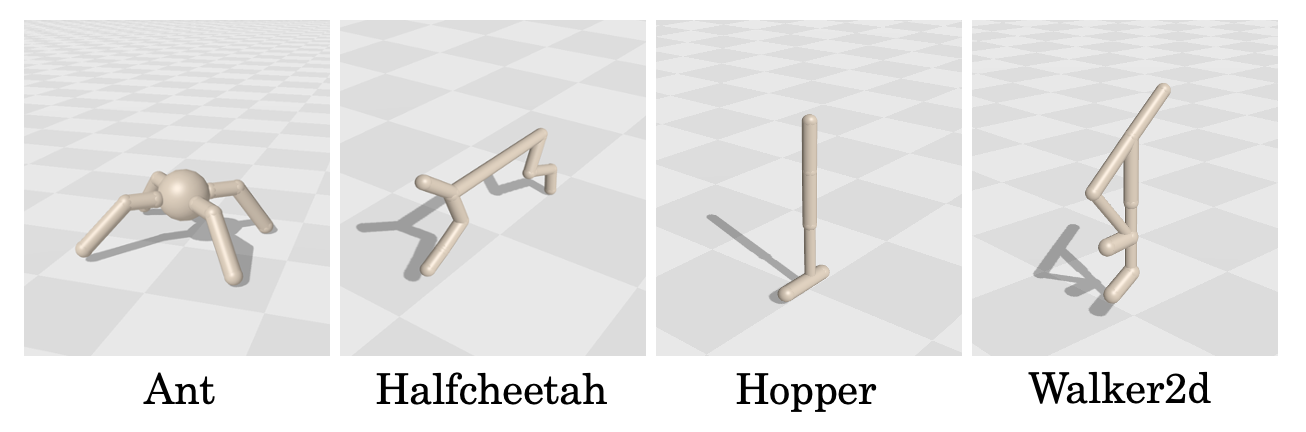}
  \caption{Brax tasks: agents aim to move forward as quickly as possible, while minimising their energy consumption.}
  \label{fig:robots}
\end{figure}

\begin{figure*}[ht]
  \centering
  \includegraphics[width=0.9\textwidth]{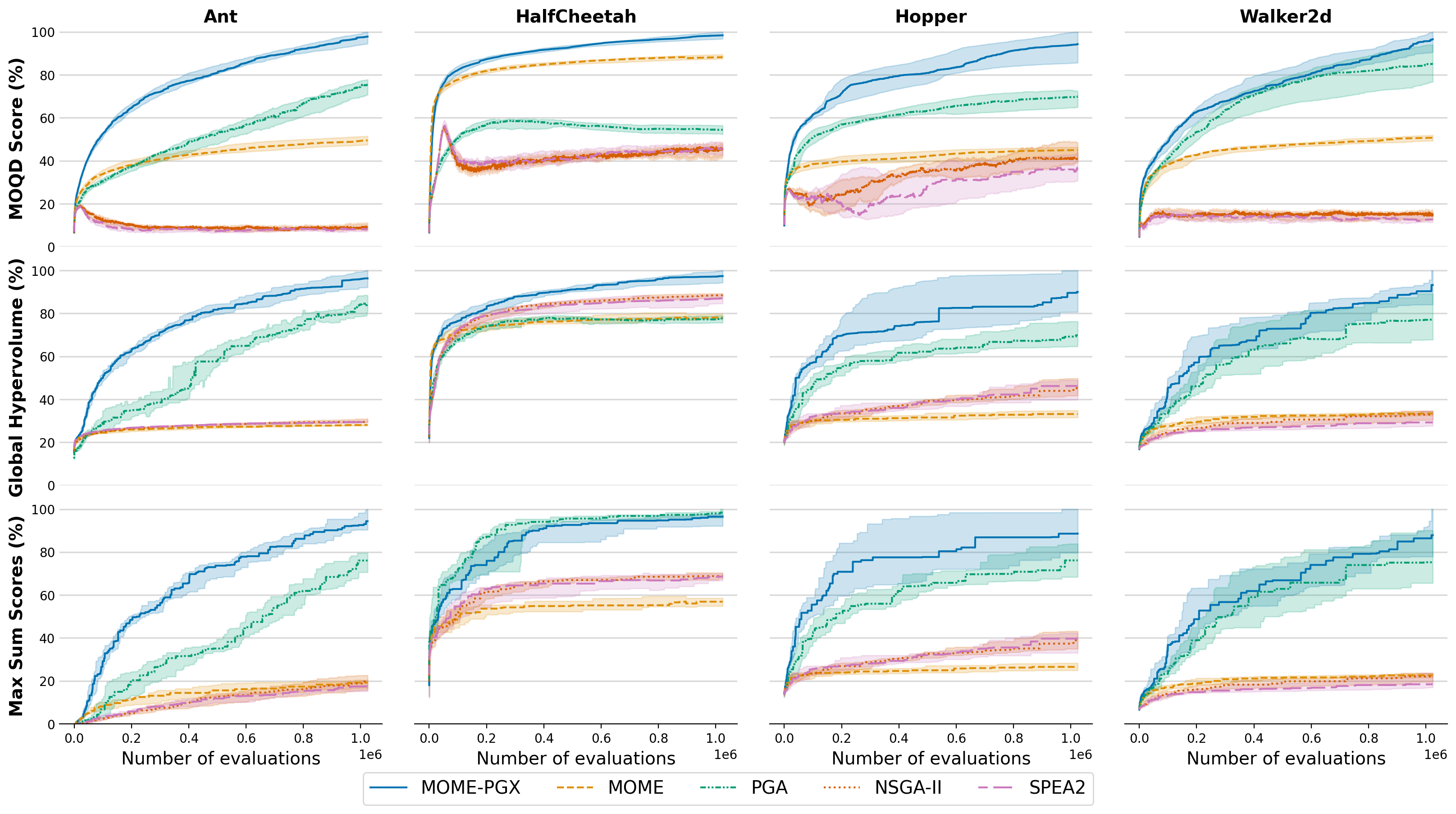}
      \caption{Performance of \pgx and all other baseline algorithms for each of the tasks. The curves show the median performance across \replications replications and the shaded regions show the inter-quartile range.
      }
    \label{fig:results}
\end{figure*}

We evaluate our approach on continuous control robotic tasks using the Brax suite \cite{brax}.
We choose four robot morphologies from the suite: the Ant, HalfCheetah, Hopper and Walker2d, visualised in Figure \ref{fig:robots}.
The solutions of the task correspond to the parameters of closed-loop neural-network controllers that control the torque commands $a_t\in [-1, 1]$ of the robot at each time step $t$, based on their current state $s_t$ characterised by the robot's proprioceptive sensors. 
For each task, we used a two layer neural network each with $64$ neurons, corresponding to solution space sizes of $6472, 5766, 5123$ and $5702$ for the Ant, HalfCheetah, Hopper and Walker2d respectively.



The aim of the task is to acquire a set of controllers that enable the robot to walk forward in different gaits, as fast as possible while minimising its energy consumption over $T=1000$ timesteps. 
To characterise a solution's gait, its descriptor corresponds to the proportion of time that the robot spends on each of its legs \cite{mome, pgamap, nature}:
\vspace{-10pt}\begin{equation}
    c_i(x) = \frac{1}{T} \sum_{t=1}^T c_{i,t} \,\,\, \text{for}\,\,i=1, ..., L\,,
\end{equation}

where $L$ is the number of legs of the robot and $c_{i,t}=1$ when the leg is in contact with the floor at time step $t$ and $0$ otherwise.
The energy consumption $f_1$ and forward motion $f_2$, are taken as:

\begin{equation}
    \begin{cases}
    f_1 = - \sum_{t=1}^T ||a_t||_2 \,,
    \\[1.2em]
    f_2 = \sum_{t=1}^T \frac{x_t - x_{t-1}}{\delta t}\,,
    \end{cases}
    \label{eqn:objs}
\end{equation}

 where $|| \cdot ||_2$ denotes the Euclidean norm, $x_t$ denotes the robot's center of gravity position at time step $t$ and $\delta t$ denotes the length of one time-step. 
 Following the implementation of the Brax authors, we also add a weighting coefficient to $f_1$, which is different for each task, to satisfy constraints on the robot's centre of gravity.

\subsection{Evaluation Metrics}

We evaluate the performance of algorithms based on four metrics:\\

\vspace{-10pt}
\textit{1)} \textbf{\moqdscore}. Given the set of all solutions in the current archive $\mathcal{A}$ and a descriptor space $\mathcal{S}$ that has been tessellated into $k$ cells $\mathcal{S}_i$, the \moqdscore is the sum of hypervolumes of Pareto fronts from each cell in the \mapelites grid \cite{mome}:
\vspace{-5pt}\begin{equation}
 \sum_{i=1}^{k} \Xi(\mathcal{P}_i), \,\,\text{where} \,\,\forall i, \mathcal{P}_i = \mathcal{P}(x \in \mathcal{A}|c(x)\in S_i)\,.
 \label{eqn:moqd}
\end{equation}

Analogous to the \qdscore in the mono-objective case \cite{qdunifying}, this metric aims to reflect both the coverage of the descriptor space and the performance of the solutions.\\

\vspace{-10pt}
\textit{2)} \textbf{\globalhypscore}.
We also evaluate performance via the global hypervolume. 
This is the hypervolume of the Pareto front formed from all solutions in the archive, equivalent to the standard Pareto front of canonical \mo algorithms \cite{mome}. 
This metric allows us to compare the best set of possible trade-offs of the objectives that we can achieve regardless of a solution's descriptor.\\

\vspace{-10pt}
\textit{3)} \textbf{\maxsumscores}. We also compare the performance of solutions using the maximum sum of scores of the objective functions. This allows direct comparison with traditional \qd algorithms, which use the sum of rewards as a single objective. \\

\vspace{-10pt}
\textit{4)} \textbf{\coverage}. The \coverage is the proportion of cells of the archive that are non-empty, reflecting the diversity of solutions in the descriptor space. This metric is reported in Appendix A as all \qd methods achieve a similar coverage.

\subsection{Baselines}
We compare \pgx to the following four baselines: 1) \mome, 2) \pga, 3) \nsga, 4) \spea.
Since \pga is not explicitly a multi-objective algorithm, we sum the two objectives in Equation \ref{eqn:objs} to form a single objective. 
Additionally, we note that using the same number of cells for \pga would not be a fair comparison. 
For example, when we use a \mapelites grid with $k$ cells each with a maximum Pareto front length of $P$, this would lead to a maximum population size of $k \times P$ in \pgx, but only a maximum population size of $k$ in \pga. 
Therefore, we use a grid tessellated into $k \times P$ cells for \pga.
By the same logic, we use a population size of $k \times P$ for \nsga and \spea. 

On the other hand, having a different number of cells for \pga and \pgx also makes comparison of the \moqdscore and \coverage unfair. 
Moreover, neither \nsga or \spea even use a tessellated grid to maintain their populations.
Therefore, for these baselines, we also maintain a passive archive with the same number of cells and maximum Pareto front length as \mome and \pgx.
At each iteration we take all of the solutions from the \pga, \nsga and \spea archives and use them to fill the passive archives via the normal Pareto front addition rules.
All metrics that we report for comparison are then calculated from these passive archives. 
Since the passive archives do not interact within the main algorithmic loop, the behaviour of the baselines remains unchanged.

\subsection{Experiment designs}
We run each experiment for the same total budget of $4000$ iterations with a batch-size of $256$, corresponding to a total of 1,024,000 evaluations. For \mome and \pgx, we use a CVT tessellation \cite{cvt} with $128$ centroids and a maximum Pareto front length of $50$ in each centroid. Accordingly, \pga uses a CVT tessellation of $128 \times 50 = 6400$ centroids, while \nsga and \spea have population sizes of $6400$. For all experiments, we chose the \ga variation operator to be Iso+LineDD operator \cite{vassiliades2018discovering} with $\sigma_1 = 0.005$ and $\sigma_2 = 0.05$. The reference points were chosen to be the empirically observed minima of the objective functions for of the each environments, as given in Appendix B, and were kept the same for each of the experiments. The hyperparameters used for neural networks in \pga, \pgx and the ablations were the same for each experiment and are provided in Appendix C.
Each experiment was repeated for \replications different seeds.

\begin{figure}[h]
  \centering
  \includegraphics[width=\linewidth]{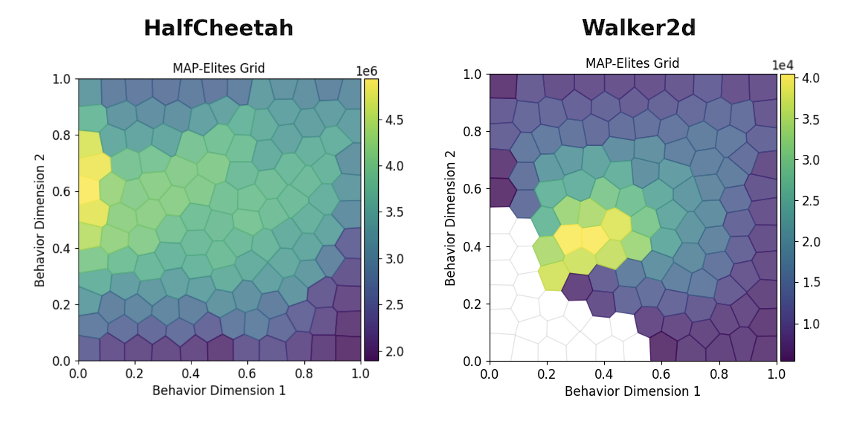}
  \caption{The final repertoires obtained from one optimisation run of \pgx, colour-coded by hypervolume. Only tasks with two-dimensional descriptors are visualised.}
  \label{fig:repertoires}
\end{figure}

\begin{figure}[h]
  \centering
  \includegraphics[width=\linewidth]{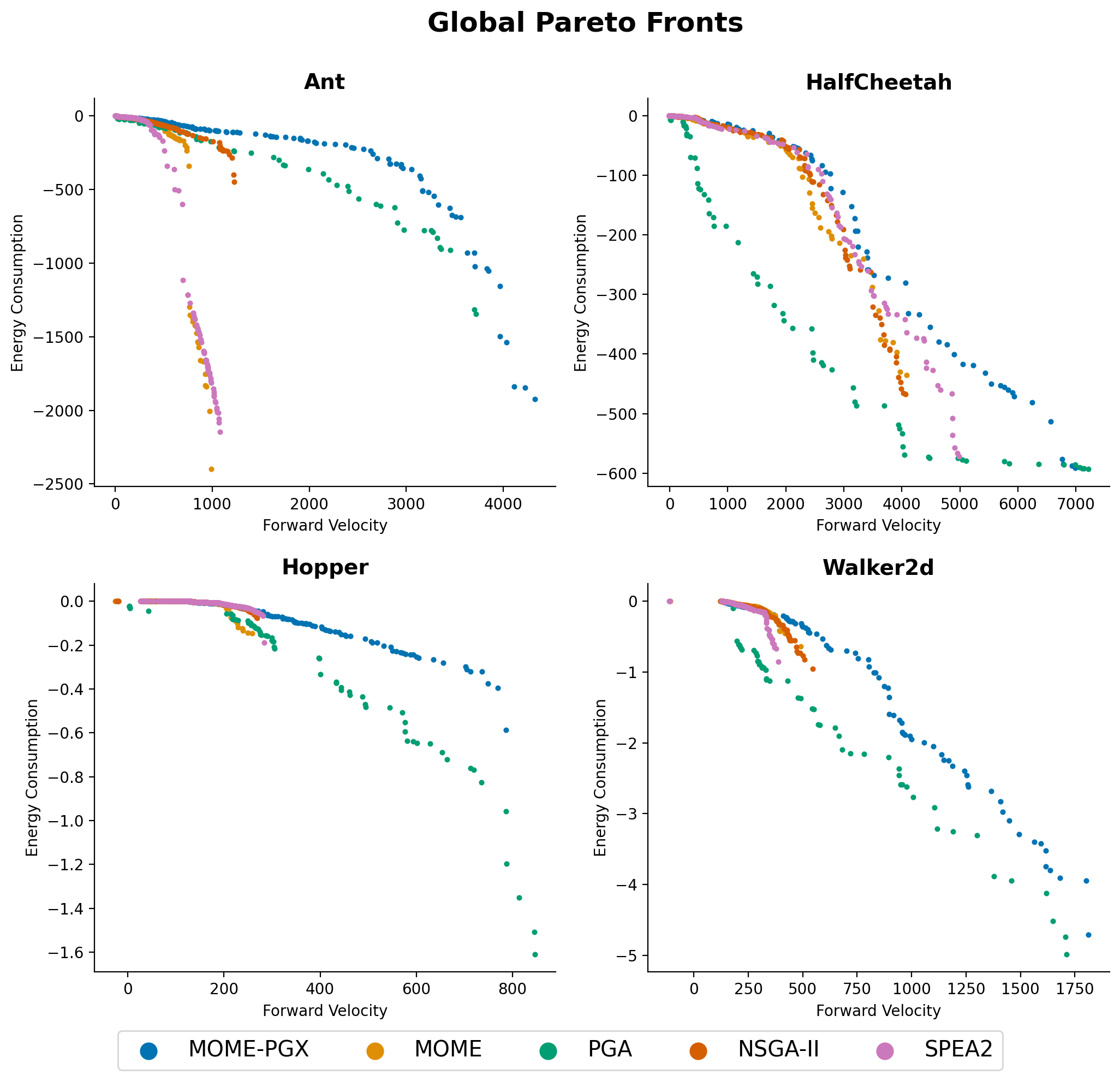}
  \caption{The global Pareto fronts obtained from one optimisation run for \pgx and all other baseline algorithms.}
  \label{fig:pfresults}
\end{figure}

\begin{figure*}[h]
  \centering
  \includegraphics[width=0.9\textwidth]{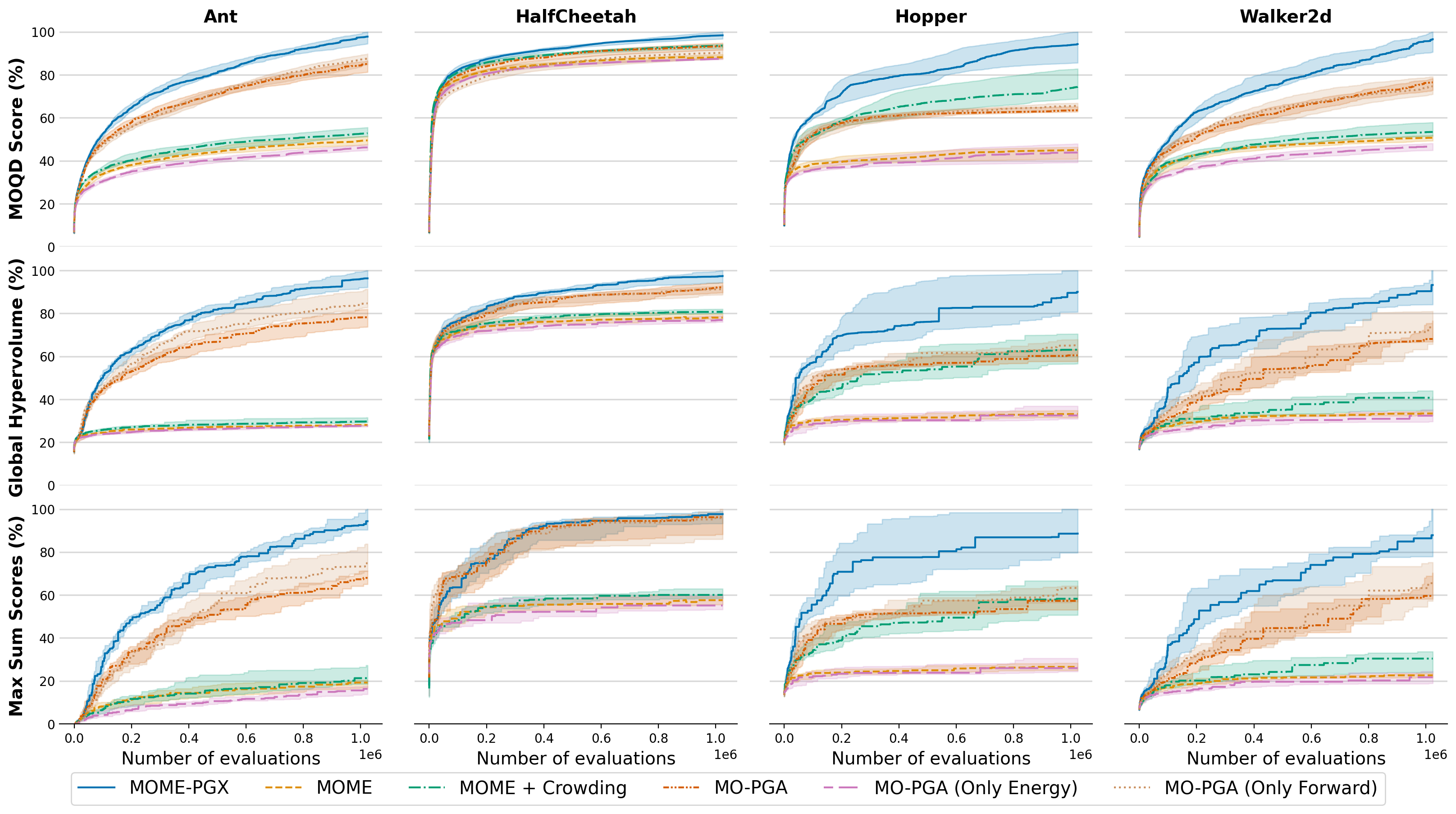}
  \caption{Performance of \pgx and all ablation algorithms for each of the tasks. The curves show the median performance across \replications replications and the shaded regions show the inter-quartile range.}
  \label{fig:ablations}
\end{figure*}

\vspace{-15pt}
\subsection{Experimental Results}

The results of our experiments are displayed in Figure \ref{fig:results}.
We also report the p-values obtained from a Wilcoxon signed-rank test \cite{wilcoxon1992individual} with a Holm-Bonferroni correction \cite{holm1979simple}, under the null hypothesis that the \pgx results come from the same distribution as the other baselines. 
The key conclusion is that \pgx outperforms or matches the performance of all other baselines, across all metrics in each of the environments.
 
The results demonstrate that \pgx achieves a higher \moqdscore in all tasks (p $<$ \moqdpvalue  in every experiment), indicating that it is better at finding high-performing Pareto fronts that span the descriptor space compared to other baselines.
This performance is highlighted by the final archives visualised in Figure \ref{fig:repertoires}.
Notably, Figure \ref{fig:results} shows that in the challenging Ant and Hopper tasks, the \moqdscore of \pgx is double that of \mome, \nsga and \spea. Across all tasks, \pgx requires between 4.3 and 42 times fewer evaluations than \mome to reach the same \moqdscore. 

\pgx also outperforms all other baselines on the \globalhypscore metric (p $<$ \globalhyppvalue  in every experiment) in all of the tasks, meaning that it is able to produce the best set of possible trade-offs across each of the objectives. 
Interestingly, \pga also outperformed the other \mo baselines, despite the fact that it does not actively seek a Pareto front of solutions.
The strong performance of \pgx and \pga highlights the value of gradient-based approaches for tasks with large search spaces. 
Indeed, examining the global Pareto fronts attained by each of the algorithms in Figure \ref{fig:pfresults}, we see that \pgx and \pga were able to find solutions that were high-performing across both of the objectives but \mome, \spea and \nsga often failed to find solutions with high forward velocity. 
We also note that, although \pga does explicitly include energy consumption within its reward, \pgx finds many solutions that achieve the same forward velocity as solutions from \pga, but at a lower energy cost. 
This is most likely due to the difference in scales of each of the objectives, meaning that the mono-objective fitness used in \pga is predominantly composed of the forward velocity reward and does not sufficiently prioritise energy consumption.
However, this highlights one problem with using such a scalarised reward, which is that finding the coefficients with which to weight each objective can prove difficult.   

Finally, we note that \pgx also demonstrates excellent mono-objective performance via the \maxsumscores metric.
\pgx outperforms all other \mo baselines (p $<$ \maxscorespvalue in every experiment). 
Moreover, \pgx outperformed \pga in the Ant task (p $=$ \psumscoresant) and there was no statistically significant difference between \pgx and \pga on the other three tasks.
This result is particularly interesting because while \pga explicitly optimises for the sum of rewards, \pgx does not. 
This suggests that optimising over each of the objectives separately could help to provide better stepping stone solutions than those found in \pga. 

\subsection{Ablation Study}
We also compare \pgx to a series of ablations.
The first, \mopga, is the same as \mome but with \pg variations for each objective. 
The second, \mopgaforward, is the same as \mopga but only uses \pg variations for the forward velocity.
Similarly, \mopgaenergy, is also the same as \mopga but only uses \pg variations for the energy consumption.
Finally, \momecrowding uses no policy gradients and simply extends \mome to include crowding-based sampling and addition.

The first key observation from the ablation study is that \pgx outperforms all ablation algorithms in the Ant, Hopper and Walker2d tasks (p $<$ \pscoreablations in all experiments).
This highlights that neither the crowding-based mechanism nor the policy gradient variation operator is solely responsible for the strong performance of \pgx but rather a combination of the two.
Interestingly, the value each of the \pgx components varies according to the task. 
For example, in the Hopper task, \momecrowding has a higher \moqdscore than \mopga (p $= 4.8\times 10^{-4}$) but the reverse is true in the Ant and Walker2d tasks (p $= 6.1\times 10^{-5}$ on both tasks). 

The second key observation is that \mopgaforward outperforms \mopgaenergy in the Ant, Hopper and Walker2d tasks (p $<$ \pscoreablations in all experiments), demonstrating that policy gradient mutations across different objectives are not equally valuable. 
In fact, \mopgaforward led to a higher a \globalhypscore and \maxsumscores than \mopga in the Ant task, and there was no statistical difference between \mopga and \mopgaforward on the \moqdscore in the Ant, Hopper and Walker2d tasks.
This can be explained by the fact that the objectives are not equally straightforward to optimise. 
Specifically, we note that low-energy solutions can be found relatively straightforwardly through \ga search mechanisms alone, as demonstrated by \mome, \spea and \nsga performances.
This lessens the need to use \pg mutations to directly minimise the energy cost and weakens the benefit of these mutations for the overall performance of the algorithm.
Despite this, since this result is likely to be highly task-specific, we still propose that \pgx uses mutations from all objective functions.
However, we note that an interesting future line of work would be to dynamically adjust the proportion of solutions mutated according to each objective \cite{multi-emitter}. 

\vspace{-10pt}\section{Conclusion}
In this work, we introduced \pgx, a new \moqd algorithm that demonstrably succeeds at  generating large collections of diverse and high-performing solutions, across multiple objective functions. 
Our experimental analysis shows that \pgx improves the sample-efficiency of \mome in tasks that can be formalised as \mdps. 
It does this by using \pg variation operators to accelerate the search process and crowding-based mechanisms to maintain exploration over the objective space. 
In future lines of work, we would like to explore \moqd approaches in tasks with many more objectives.

\vspace{-10pt}\section*{Acknowledgements}
This work was supported by PhD scholarship funding for Hannah from InstaDeep.

\bibliographystyle{ACM-Reference-Format}
\bibliography{main}

\newpage

\section*{Appendices}

\appendix

\section{Coverage Results}\label{app:coverage}

\begin{figure}[H]
  \centering
  \includegraphics[width=0.85\linewidth]{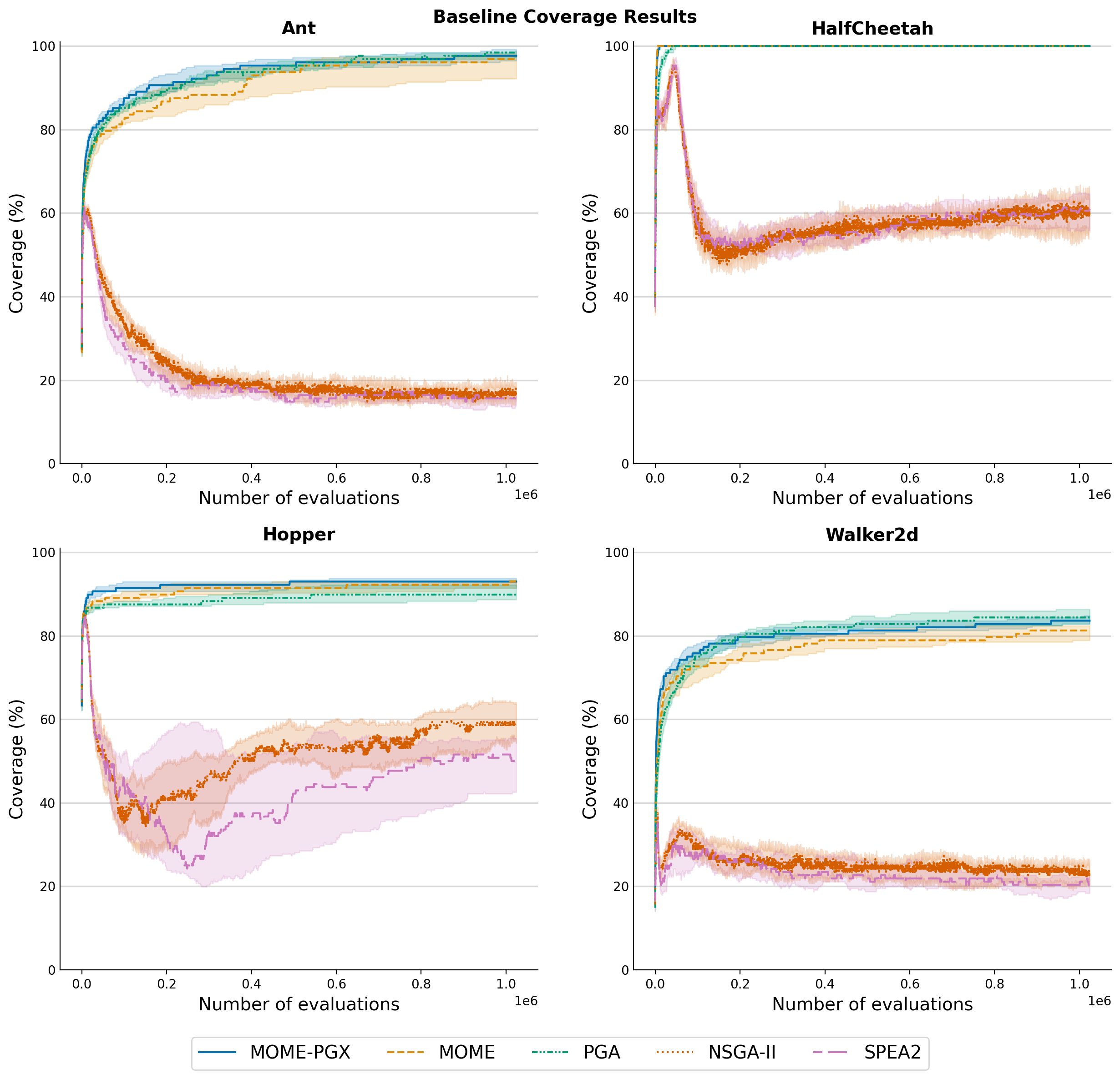}
  \caption{Coverage performance of \pgx and all baseline algorithms for each of the tasks. The curves show the median performance across \replications replications and the shaded regions show the inter-quartile range.}
  \label{fig:coverage}
\end{figure}

\begin{figure}[H]
  \centering
  \includegraphics[width=0.85\linewidth]{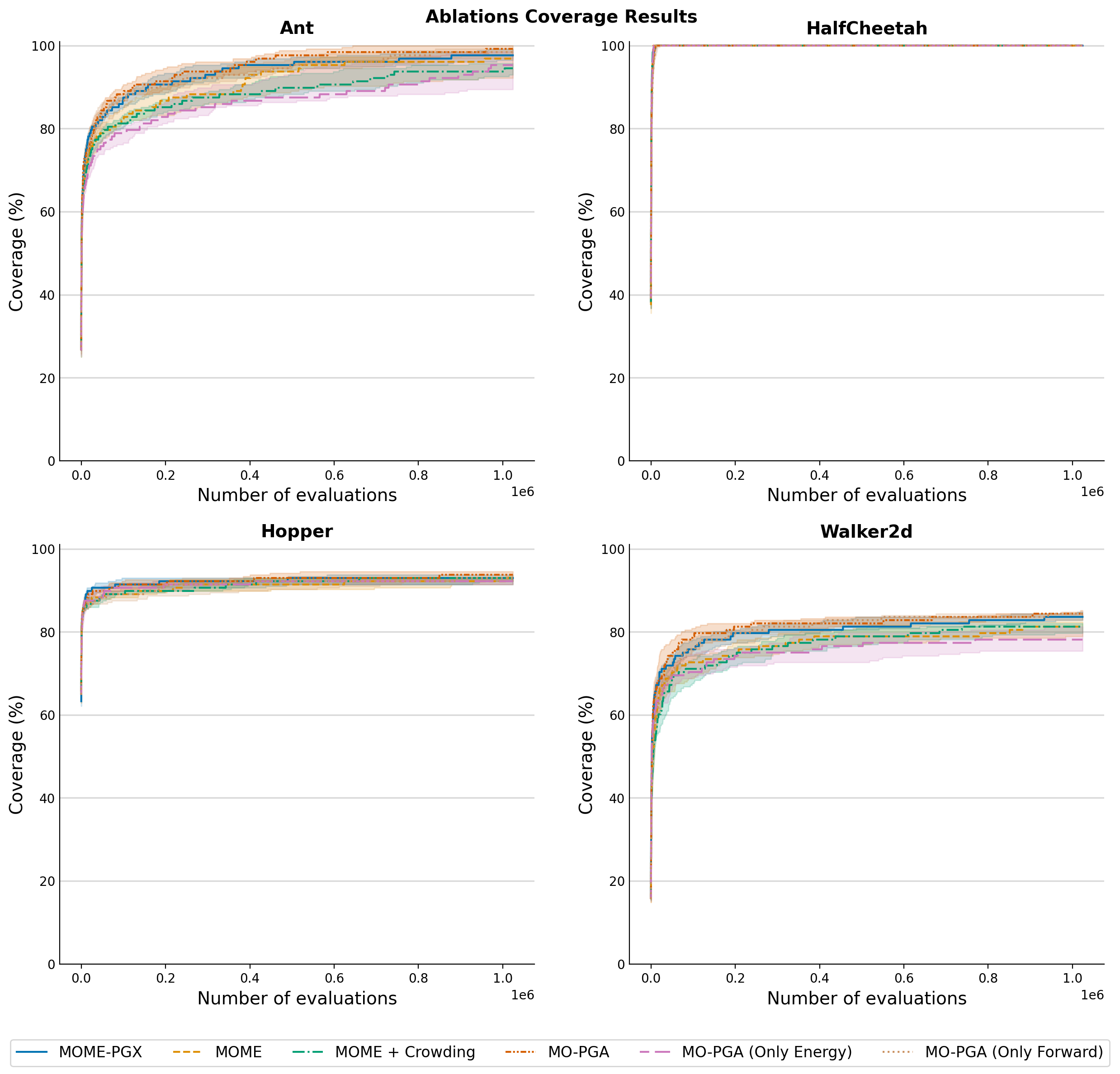}
  \caption{Coverage performance of \pgx and all ablation algorithms for each of the tasks. The curves show the median performance across \replications replications and the shaded regions show the inter-quartile range.}
  \label{fig:ablations_coverage}
\end{figure}

Figure \ref{fig:coverage} and Figure \ref{fig:ablations_coverage} show the coverage achieved across \replications replications in each of the experiments. 
The results show that all \qd algorithms manage to find solutions which span the entire descriptor space. 
Since \spea and \nsga do not seek diversity over the descriptor space, they achieve comparably poor coverage.

\section{Hypervolume Reference Points} \label{app:refpoints}
Table \ref{tab:refpoints} presents the reference points used to calculate the hypervolume metrics in each of the tasks. 
The same reference points were used for all of the experiments.

\begin{table}[h!]
    \centering
        \caption{Reference points}
    \begin{tabular}{l | l }
        \toprule
    Ant & [-350, -4500] \\
    HalfCheetah & [-2000, -800] \\
    Hopper & [-50, -2] \\
    Walker2d & [-210, -15] \\
        \bottomrule
    \end{tabular}      	
    \label{tab:refpoints}
\end{table}

\section{Policy Gradient Hyperparameters}\label{app:pghyperparams}

Table \ref{tab:pghyps} presents all of the policy gradient hyperparameters that are used for our algorithms.
All hyperparameters were kept the same for each task and for all algorithms which used PG variations.

\begin{table}[h!]
    \centering
        \caption{Policy Gradient Network Hyperparameters}
    \begin{tabular}{l | l }
        \toprule
    Replay buffer size & 1,000,000 \\
    Critic training batch size & 256\\
    Critic layer hidden sizes & $[256, 256]$\\
    Critic learning rate &  $3\times 10^{-4}$\\
    Actor learning rate &  $3\times 10^{-4}$\\
    Policy learning rate &  $1\times 10^{-3}$\\
    Number of critic training steps & 300\\
    Number of policy gradient training steps & 100\\
    Policy noise & 0.2\\
    Noise clip & 0.2\\
    Discount factor & 0.99\\
    Soft $\tau$-update proportion &  0.005\\
    Policy delay & 2\\
        \bottomrule
    \end{tabular}      	
    \label{tab:pghyps}
\end{table}

\end{document}